\documentclass[]{article}
\usepackage{graphicx}
\usepackage{authblk}
\usepackage[maxnames=1,style=numeric,url=false,sorting=none,natbib=true,]{biblatex}
\usepackage{hyperref}
\usepackage[utf8]{inputenc}
\usepackage{siunitx}
\usepackage{wrapfig}
\usepackage{pifont,mdframed}

\newenvironment{warning}
  {\par\begin{mdframed}[linewidth=2pt,linecolor=red]%
    \begin{list}{}{\leftmargin=1cm
                   \labelwidth=\leftmargin}\item[\Large\ding{43}]}
  {\end{list}\end{mdframed}\par}

\begin{document}
\title{Brain-Inspired Hardware for Artificial Intelligence: Accelerated Learning in a Physical-Model Spiking Neural Network}

\author[1,*]{Timo Wunderlich}
\author[1,**]{Akos F. Kungl}
\author[1]{Eric Müller}
\author[1]{Johannes Schemmel}
\author[1,2]{Mihai A. Petrovici}
\affil[1]{\small Kirchhoff Institute for Physics, Heidelberg University, Heidelberg, Germany}
\affil[2]{Department of Physiology, University of Bern, Bern, Switzerland}
\affil[*]{timo.wunderlich@kip.uni-heidelberg.de}
\affil[**]{fkungl@kip.uni-heidelberg.de}
\begin{warning}
This article has been published as part of the Lecture Notes in Computer Science book series (LNCS, volume 11727). The final authenticated version is
available online at  \url{https://doi.org/10.1007/978-3-030-30487-4_10}.\\
Please cite it as \emph{Wunderlich et. al, 2019. Brain-Inspired Hardware for Artificial Intelligence: Accelerated Learning in a Physical-Model Spiking Neural Network. In: Tetko I., Kůrková V., Karpov P., Theis F. (eds) Artificial Neural Networks and Machine Learning – ICANN 2019: Theoretical Neural Computation. ICANN 2019. Lecture Notes in Computer Science, vol 11727. Springer, Cham. doi: 10.1007/978-3-030-30487-4\_10.}.
\end{warning}
{\let\newpage\relax\maketitle}

\begin{abstract}
Future developments in artificial intelligence will profit from the existence of novel, non-traditional substrates for brain-inspired computing.
Neuromorphic computers aim to provide such a substrate that reproduces the brain's capabilities in terms of adaptive, low-power information processing.
We present results from a prototype chip of the BrainScaleS-2 mixed-signal neuromorphic system that adopts a physical-model approach with a 1000-fold acceleration of spiking neural network dynamics relative to biological real time.
Using the embedded plasticity processor, we both simulate the Pong arcade video game and implement a local plasticity rule that enables reinforcement learning, allowing the on-chip neural network to learn to play the game.
The experiment demonstrates key aspects of the employed approach, such as accelerated and flexible learning, high energy efficiency and resilience to noise.
\end{abstract}
\pagebreak
\section{Introduction}
Many breakthrough advances in artificial intelligence incorporate methods and algorithms that are inspired by the brain.
For instance, the artificial neural networks employed in deep learning are inspired by the architecture of biological neural networks \citep{Hassabis2017Neuroscience-InspiredIntelligence}.
Very often, however, these brain-inspired algorithms are run on classical von Neumann devices that instantiate a computational architecture remarkably different from the one of the brain.
It is therefore a widely held view that the future of artificial intelligence will depend critically on the deployment of novel computational substrates \citep{Dean2018ARevolution}.
Neuromorphic computers represent an attempt to move beyond brain-inspired software by building hardware that structurally and functionally mimics the brain \citep{Schuman2017AHardware}.

In this work, we use a prototype of the BrainScaleS-2 (BSS2) neuromorphic system \citep{Friedmann2017DemonstratingSystem}.
The employed physical-model approach enables the accelerated (1000-fold with respect to biology) and energy-efficient emulation of spiking neural networks (SNNs).
Beyond SNN emulation, the system contains an embedded plasticity processing unit (PPU) that provides facilities for the flexible implementation of learning rules.
Our prototype chip (see Fig. \ref{fig:prototype} A) contains 32 physical-model neurons with 32 synapses each, totalling 1024 synapses.
The neurons are an electronic circuit implementation of the leaky integrate-and-fire neuron model.
We use this prototype to demonstrate key advantages of our employed approach, such as the 1000-fold speed-up of neuronal dynamics, on-chip learning, high energy-efficiency and robustness to noise in a closed-loop reinforcement learning experiment.

\section{Experiment}
\begin{figure}
    \centering
    \includegraphics[width=.9\textwidth]{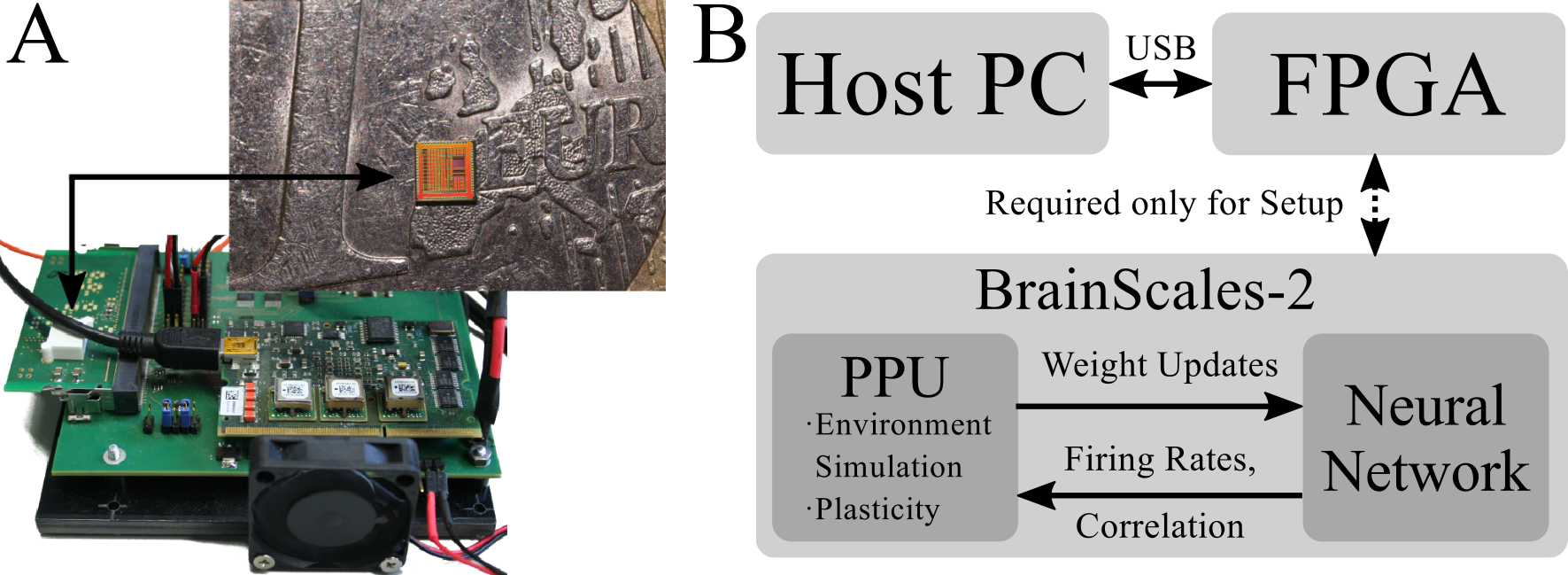}
    \caption{A: Prototype chip and evaluation board.
    B: Schematic of experimental setup. The chip runs the experiment fully autonomously, with the PPU simulating the virtual environment (Pong) and calculating reward-based weight updates that are applied to the on-chip analog neural network.
    The neural network receives the discretized ball position as input and outputs the target paddle position.
    B taken from \cite{Wunderlich2019DemonstratingStudy}.}
    \label{fig:prototype}
\end{figure}
The experiment represents the first demonstration of on-chip closed-loop learning in an accelerated physical-model neuromorphic system \citep{Wunderlich2019DemonstratingStudy}.
It takes place on the chip fully autonomously, with external communication only required for initial configuration (see Fig. \ref{fig:prototype} B).
We use the embedded plasticity processor both to simulate a simplified version of the Pong video game (opponent is a solid wall) and to implement a reward-modulated spike-timing-dependent plasticity (R-STDP) learning rule \citep{Fremaux2015NeuromodulatedRules.} of the form $\Delta w_{ij} \propto (R-\bar R)e_{ij}$, where $R$ is the reward, $\bar R$ is a running average of the reward and $e_{ij}$ is an STDP-like eligibility trace.
Each synapse locally records the STDP-like eligibility trace and stores it as an analog value (a voltage), to be digitized and used by the plasticity processor \citep{Friedmann2017DemonstratingSystem}.

The two-layer neural network receives the ball position along one axis as input and dictates the target paddle position, to which the paddle moves with constant velocity, using the neuronal firing rates.
It receives reward depending on its aiming accuracy (i.e., how close it aims the paddle to the center of the ball), with $R=1$ for perfect aiming, $R=0$ for not aiming under the ball and graded steps in between.
By correlating reward and synaptic activity via the given learning rule, the SNN on the chip learns to trace the ball with high fidelity.
\section{Results}

\begin{figure}[h!]
    \centering
    \includegraphics[width=10cm]{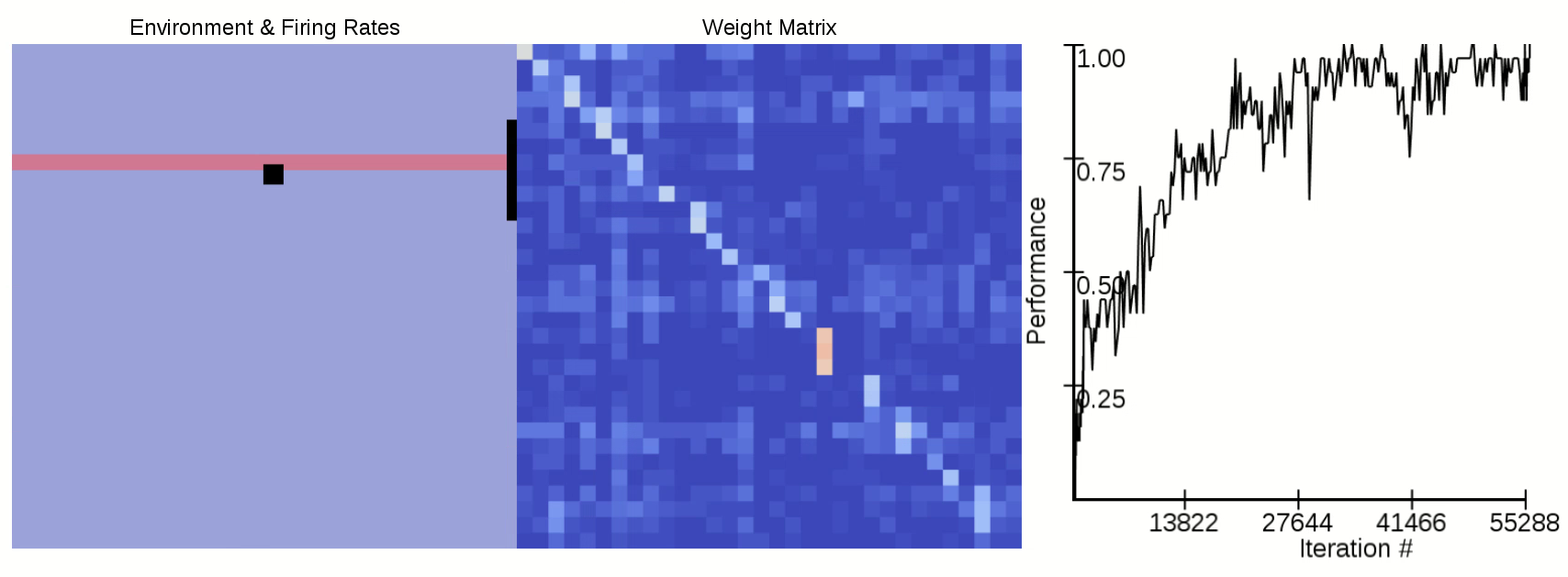}
    \caption{Screenshot of experiment interface for live demonstration.
    Left: Playing field and color-coded neuronal firing rates.
    Middle: Color-coded synaptic weight matrix.
    Right: Performance in playing Pong.
    }
    \label{fig:demo}
\end{figure}

A screen recording of a live demonstration of the experiment (see Fig. \ref{fig:demo}) is available at \url{https://www.youtube.com/watch?v=LW0Y5SSIQU4}.
The recording allows the viewer to follow the game dynamics, neuronal firing rates, synaptic weight dynamics and learning progress.
The learning progress is quantified by measuring the relative number of ball positions for which the paddle is able to catch the ball.
The learned weight matrix is diagonally dominant: this expresses a correct mapping of states to actions in the reinforcement learning paradigm.

Importantly, neuronal firing rates vary from trial to trial due to noise in the analog chip components.
Used appropriately, this can become an asset rather than a nuisance: in our reinforcement learning scenario, such variability endows the neural network with the ability to explore the action space and thereby with a necessary prerequisite for trial-and-error learning.
We also found that neuronal parameter variability due to fixed-pattern noise, which is inevitable in
\begin{wrapfigure}{l}{.4\textwidth}
\vspace{-0cm}
\begin{center}
    \includegraphics[clip, trim=.5cm .9cm .3cm .7cm, width=.4\textwidth]{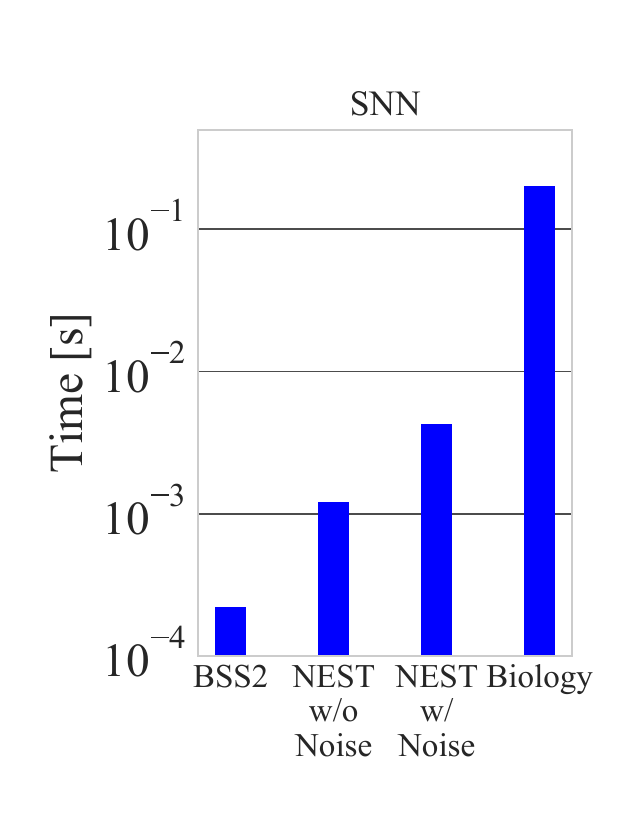}
    \caption{Comparison of experiment durations.  w/o: without. w/: with.
    Adapted from \cite{Wunderlich2019DemonstratingStudy}.}
    \label{fig:time}
\end{center}
\vspace{-0cm}
\end{wrapfigure}
analog neuromorphic hardware, is implicitly compensated by the chosen learning paradigm, leading to a correlation of learned weights and neuronal properties \citep{Wunderlich2019DemonstratingStudy}.

The accelerated nature of our substrate represents a key advantage.
We found that a software simulation (NEST v2.14.0) on an Intel processor (i7-4771), when considering only the numerical state propagation, is at least an order of magnitude slower than our neuromorphic emulation (see Fig. \ref{fig:time} and \cite{Wunderlich2019DemonstratingStudy}).
Besides this, the emulation on our prototype is at least 1000 times more energy-efficient than the software simulation (\SI{23}{\micro\joule} vs. \SI{106}{\milli\joule} per iteration).
This evinces the considerable benefit of using the BSS2 platform for emulating spiking networks and hints towards its decisive advantages when scaling the emulated networks to larger sizes.

\section{Conclusions}
These experiments demonstrate, for the first time, functional on-chip closed-loop learning on an accelerated physical-model neuromorphic system.
The employed approach carries the potential to both enable researchers with the ability to investigate learning processes with a 1000-fold speed-up and to enable novel, energy-efficient and fast solutions for brain-inspired edge computing.
While digital neuromorphic solutions and supercomputers generally achieve at most real-time simulation speed in large-scale neural networks \citep{Jordan2018ExtremelyComputers,vanAlbada2018PerformanceModel,Mikaitis2018NeuromodulatedSystem}, the speed-up of BrainScaleS-2 is independent of network size and will become a critical asset in future work on the full-scale BrainScaleS-2 system.
\pagebreak
\section*{Funding}
This research was supported by the EU 7th Framework Program under grant agreements 269921 (BrainScaleS), 243914 (Brain-i-Nets), 604102 (Human Brain Project), the Horizon 2020 Framework Program under grant agreements 720270 and 785907 (Human Brain Project) and the Manfred Stärk Foundation.

\printbibliography
\end{document}